# Towards understanding and modelling office daily life

Michele Bezzi      Robin Groenevelt

Accenture Technology Park,
Sophia Antipolis, F-06902, France

**ABSTRACT**

Measuring and modeling human behavior is a very complex task. In this paper we present our initial thoughts on modeling and automatic recognition of some human activities in an office. We argue that to successfully model human activities, we need to consider both individual behavior and group dynamics. To demonstrate these theoretical approaches, we introduce an experimental system for analyzing everyday activity in our office.

**Keywords**
Probabilistic data, office activities, information theory; social networks

**INTRODUCTION**

People and businesses have a natural interest in studying human behavior patterns. This can come forth from security concerns, to offer improved health care of individuals, to increase and monitor the performance of people, to understanding how customers behave, to optimize organizational structure, or to improve communications among groups of people.

Unfortunately finding reliable mathematical descriptions of human activities is extremely difficult even in well structured environments such as an office. The complexity of human behavior arises from the interaction between two main levels: *individual behavior* and *group dynamics*.

Individuals are *per se* complex entities: their actions depend not only on the sensory context, but also on various hard-to-measure factors such as past personal history, attention, attitudes, experiences, and emotions. To investigate these complex patterns of activity we need to consider the actual context and the context history. For example, collecting sensory information for long periods (e.g. months) we can search for frequent recurrent patterns of activity (habits), and, accordingly, create a statistical model of people's daily activities. Deviations from this baseline may indicate a change from routine activity. Due to the high variability that characterizes human behavior, this process generates a huge number of patterns. Similarly, the redundancy and the complex hierarchical structure of habitual behavior [1] (a complex habit may be decomposed into many simpler sub-habits) also produce a multitude of recurrent patterns. In our approach we will apply a combination of context specific knowledge and statistical methods to choose appropriate models or to select specific features of certain behaviors. The choice of the temporal and spatial scale plays also an important role, e.g. decreasing the spatial resolution (large spatial bins) may help to compensate for the inherent stochasticity in people movement patterns, but it may lead to a large information loss, as well. Again, context knowledge and physical constraints may be used to choose the appropriate temporal and spatial resolution.

An additional source of stochasticity is the presence of noise at sensor level. Sensor networks producing large quantities of (often) redundant, but noisy, data. In fact, although sensor technology is rapidly progressing, undetected events and false positive are almost always present in any sensor network. Thus to fully exploit the data we should be able to handle the intrinsic noisy nature of sensor data. In our case, data coming from multiple heterogeneous sensory sources are integrated using a Bayesian framework [2,3] that combines probabilistic and knowledge-based approaches.

On the positive side, recent advances in sensor technologies provide us a large amount of data about human behaviour in every day life. Taking advantage of these large data sets and sensor redundancy we may partly compensate for the stochasticity at the sensor and behavioral level, and improve precision and robustness of the system. Furthermore, observing real environments for long periods of time may reveal dynamics that are not evident from small-scale studies in artificial environments and for limited durations [4].

Group dynamics, often due to social interactions, are also highly complex processes. It has been found that networks of friendships or personal contacts can exhibit small world [5,6] or scale-free properties [7], i.e., there are many people with few connections and a few people with many connections. An important aspect of our study on behaviour comes forth from human physical interactions. To estimate this we will focus on the movement trajectories of people.

In this paper we present a system we are developing to detect and measure various behaviors in everyday office life. We will briefly describe our experimental environment and numerical simulations of office life, after which we will present some preliminary results related to detecting unusual activities and social connections. Finally we will discuss some potential issues when deploying such a system.

**MODELING OFFICE ACTIVITIES**

We have chosen an office environment as a test setting for various reasons. First of all, a quantitative description of various office activities may have important practical applications (e.g. assessing the quality of space organization in the office, estimating connections amongst different people/departments, safety and security). Secondly, a video-camera infrastructure is readily available in our location and the data are easily accessible. Finally, data from the camera systems can be integrated with, or replaced by, other sensors (ultra wide band tracking devices, badge readers, finger print readers) and with data extensively available in electronic form (calendars, e-mails, log files).

The actual functionality of our system will be determined using probabilistic tracking data from Accenture labs in Chicago [2,3]. This modular system provides long term recordings and probabilistic tracking. Along with real world data, we are implementing a numerical simulation of people their movements in an office analogous to the one used for collecting real world data.

**Experimental setup**
This section describes a probabilistic framework for identifying and tracking moving objects using multiple streams of sensory data (a more detailed description can be found in [2,3]).
The experimental environment is composed of an office floor at Accenture Technology Labs in Chicago. The floor is equipped with a network consisting of 30 video cameras, 90 infrared tag readers, and a biometric station for fingerprint reading.
The first step is the fusion of this raw-sensor data into a higher-level description of people's movements inside the office. People identification and tracking is performed using a Bayesian network. In short (see [3] for details), the office space is divided into 50 locations, each of them the size of a small office. This allows us to remove the variability of paths inside a room while still maintaining enough information about people their movements. Each sensor detects signals of people in its sensory field. For each person and location the signals are merged together to build the current probabilistic evidence of finding a certain person in a specific location, after which this information is integrated with the current belief of the system (originated by previous observation). The result is a sequence of matrices, one for each time step, where the probability finding a person in each location is reported.

In the second step, starting from these matrices, we derive the most likely paths for each tracked individual; these data are then analysed to find frequent patterns, appropriate statistical quantities to describe long term activities. Extracted recurrent patterns may be later identified exploiting local semantics (e.g. meetings usually take place in the meeting room) and context-knowledge (e.g. matching movement patterns with the information available from the electronic calendar).

The data acquisition system is still under development, so we have too few data currently available to find meaningful recurrent patterns. Nonetheless, to give a glimpse of the kind of statistical analysis we are interested in, we have analyzed a limited data set.

For example, we have measured the time spent in each location *x* by each single user across a number of days, *P(x)*, and for each single day, *P(x|day)*. See Figure 1. The behavior on a single day is then compared to an average day, estimating the so-called *stimulus specific information* (also called *surprise* [9]) for each day:

$$I(day) = \sum_{x \in X} P(x|day) \log_2 \frac{P(x|day)}{P(x)}$$

This quantity is large in case of *surprising* (different from the average) patterns. The main advantage of this statistical quantity is that it is *additive* (i.e. it fulfills the chain rule, as mutual information, see [9]). This allow us to easily integrate other sources of information (e.g. log-files) by simply summing the corresponding specific information.
We observe a clear peak on day 5, (Fig. 1c) indicating some unusual behavior on that day.

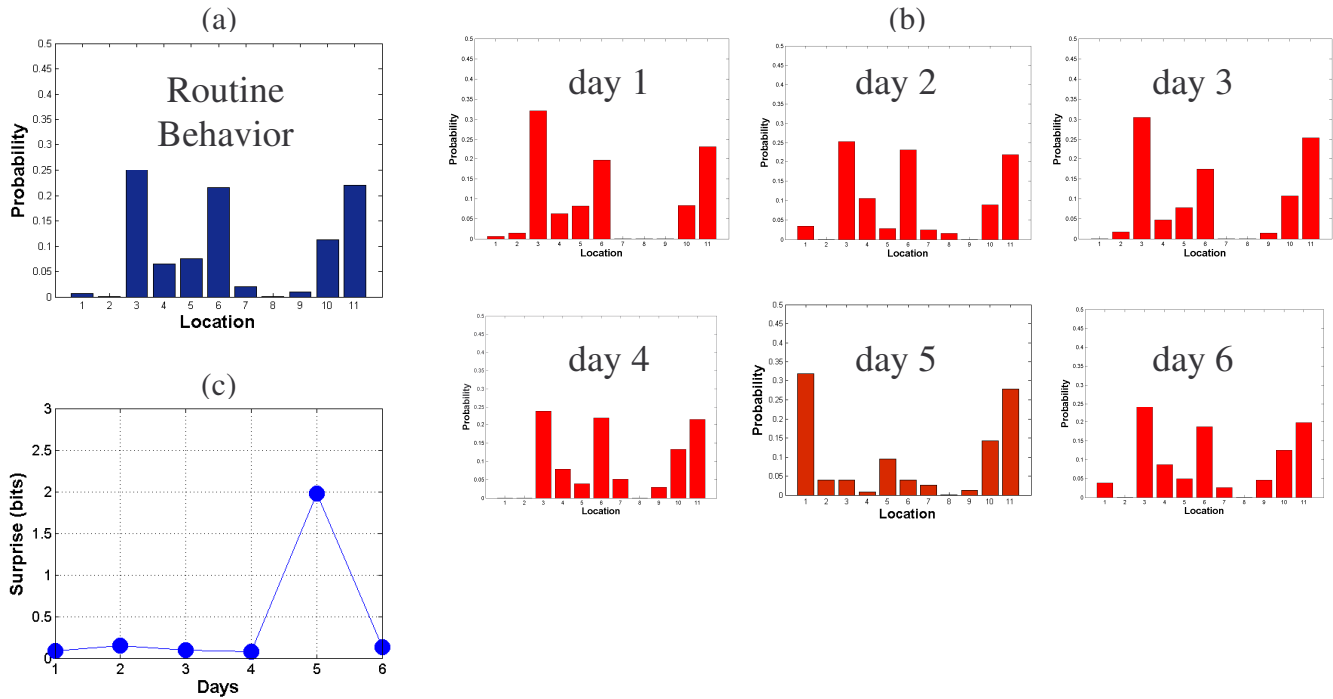

**Figure 1**. Measuring deviation from routine behavior. (a) Distribution of occupancy time across one week for one person over different office locations. (b) Distribution of occupancy time for each single day. (c) Surprise as a function of day of the week. *Surprise* quantifies the amount of mutual information we gain observing occupancy time distribution for one day (P(x|day)). Large values indicate surprising---unusual---behavior.

**Numerical Simulation**

Data collection in a real environment is a long process and it may generate privacy concern. Therefore, to freely test our algorithms and hypothesis, we have also built an agent-based simulation of people movements in the office. In its simplest version, each agent has a set of possible destinations in the office floor (with different probabilities derived from real data and from our knowledge of office life). At each time step, each agent decides to stay in the current location with a certain probability (usually large if it is in its own office) or to move to a destination sampled from a destination distribution. In this last case, it starts moving according to a specific path (usually the shortest one) with possible random fluctuations. An agent also has a personal schedule where specific events are listed (meetings, lunch time, etc ...) with a corresponding time and probability to perform that action. In addition, the probability of staying in a certain location is increased by a quantity Δp, specific for each agent, whenever it is in the presence of presence of other agents in the same location. These probabilities are derived from real data when available, and using context knowledge in the other cases.

Despite the simplicity of this model, a preliminary set of simulations show a good visual agreement with the trajectories observed in the real environment

**SOCIAL NETWORK**

Social network analysis has a long history [5], but has only recently been able to take advantage of the large use of digital communications; the properties of such networks have been extensively studied using data from emails [10] and instant-messaging [11]. In this framework, an individual's email spool file is analyzed and a list of his connections is automatically generated and displayed as a graph. Typical analyses include: number and frequency of contacts, time evolution of the network, identifying most-connected nodes, etc… . Although this approach gives interesting results and potential applications [12,13], it does not consider physical interactions and face-to-face communications that are at the basis of human behaviours.

Analyzing social network structure and its temporal evolution, we can identify networks *hubs* (key people) and infer the (possible) hierarchical structure of the group (leaders, followers), the existence of groups of interests, or potential communication gaps (conflicts) among groups. Using this analysis we may, for instance, assess the impact of change in the environment on the social structure, or the effects of team building exercise or collaboration on the personal contact network.

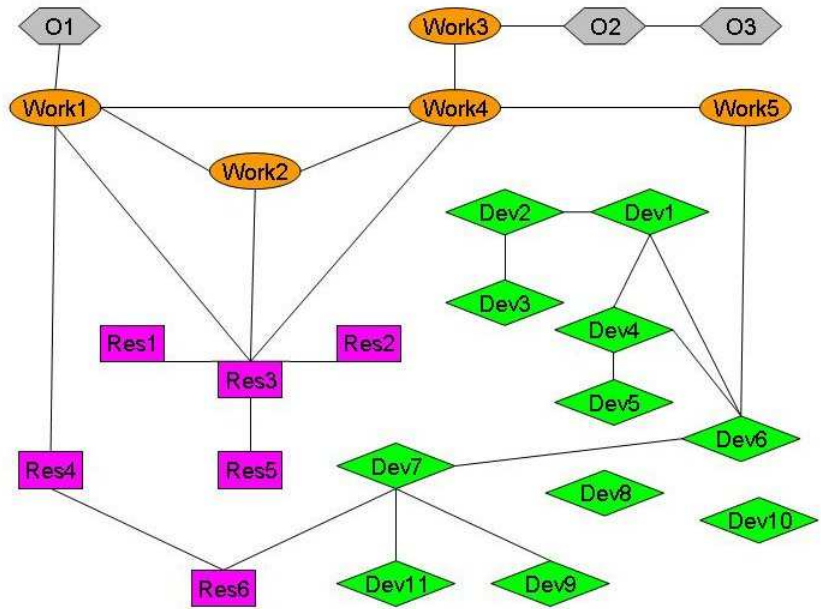

**Figure 2.** Social network of personal contacts in our labs as extracted from movement data from a preliminary test of one day of data. The different shapes represent the various departments: Research (squares), Development (diamonds), Workshops (ovals), and other people (hexagons). The real names have been replaced by labels.

In our case, tracking the movement of a group of people for long periods, we can infer the structure of the underlying social network, considering a simple *proximity* rule (i.e. two individuals share a link if they spend *enough* time in the vicinity of one another) and directionality of the interaction (i.e. going to Bob's office is different from Bob coming to visit me). In addition, we added to the system some context specific rules, e.g., people standing next to the printer do not necessarily interact, or similarly, two office mates sharing the same table. Figure 2 illustrates the social network amongst the three departments as measured in our lab (preliminary test using one day of video camera recordings).

This simple rule may lead to a large number of false positives and it also it is limited by the range of sensor network. However, we expect that in the long run and with a large number of users it may provide a reasonable first approximation of global structure of the network of interactions and of its evolution in time. This approach will be integrated with more standard methods based on electronic communications to better specify the structure of the network and to investigate the (possible) different topologies of electronic and physical social networks.

## CONCLUSIONS

Automatic recognition and prediction of human activities from sensory observations is a fast growing research field. Many technical issues are starting to be solved in laboratory settings, but there remain many technical and social obstacles for a successful deployment in real life environments. The great variability of human behavior even in rather simple activities is the main technical obstacle for automatic detection, but social aspects are not less important. Let us briefly discuss a couple of them:

Privacy is a major concern for pervasive technology acceptance. Possible solutions include a users control on personal data release, limiting the data a single party can access, encryption, data anonymization, and accepted ethical guidelines. In applications where real-time is not a requirement (as in our case for identifying social networks), the users may have full control on the data release, e.g.; receiving a weekly e-mail with the summary of events; and deciding which of them to disclose for the analysis. Even more important is finding a reasonable equilibrium point in the trade-off between privacy and benefits. In other words, users need to be provided a clear and tangible return for their privacy investment for gaining acceptance.

Artificial behavior: The use of behavior analysis for practical purpose (e.g. monitoring health status, assessing performance) may induce people to behave artificially, i.e. to behave in a non-natural way to mimic expected patterns. This is not necessarily negative, for example, if such a system is used to assess the compliance with some safety procedures, but it should be taken into account when analyzing behavioral data. We may expect this bias to decrease with an increasing user acceptance of pervasive technologies.

In conclusion, we are implementing a system for automatic analysis of some behaviors in everyday office life. Although a fully automatic system for recognition of human activities in real world situations is still far in the future, focusing on a specific context and exploiting the large availability of past and present data, we may derive a quantitative description for some of these activities, which are useful for practical purpose.

**ACKNOWLEDGMENTS**
We thank Agata Opalach for providing helpful comments on previous versions of this document. We also thank Valery Petrushin and Gang Wei for providing tracking data obtained from Accenture Technology Labs in Chicago, and Frederick Schlereth for performing the numerical simulations.